# Bio-inspired circular soft actuators for simulating defecation process of human rectum

Zebing Mao[1*], Sota Suzuki[2], Ardi Wiranata[3], Yanqiu Zheng[4], Shoko Miyagawa[5]

[1]Faculty of Engineering, Yamaguchi University, Yamaguchi, Japan
[2]School of Engineering, Tokyo Institute of Technology, Tokyo, Japan
[3]Department of Mechanical and Industrial Engineering, Universitas Gadjah Mada, Indonesia,
[4]Department of Mechanical Engineering, Ritsumeikan University, Shiga, Japan
[5]Faculty of Nursing and Medical Care, Keio University, Kanagawa, Japan

Corresponding: mao.z.aa@yamguchi-u.ac.jp

## Abstract

Soft robots have found extensive applications in the medical field, particularly in rehabilitation exercises, assisted grasping, and artificial organs. Despite significant advancements in simulating various components of the digestive system, the rectum has been largely neglected due to societal stigma. This study seeks to address this gap by developing soft circular muscle actuators (CMAs) and rectum models to replicate the defecation process. Using soft materials, both the rectum and the actuators were fabricated to enable seamless integration and attachment. We designed, fabricated, and tested three types of CMAs and compared them to the simulated results. A pneumatic system was employed to control the actuators, and simulated stool was synthesized using sodium alginate and calcium chloride. Experimental results indicated that the third type of actuator exhibited superior performance in terms of area contraction and pressure generation. The successful simulation of the defecation process highlights the potential of these soft actuators in biomedical applications, providing a foundation for further research and development in the field of soft robotics.

# 1. Introduction

Soft robotics represents a new paradigm in engineering, markedly different from traditional rigid robotics [1][2][3][4][5][6]. This innovative field has garnered considerable attention due to its ability to undergo significant deformation, exhibit exceptional softness, conform to irregular shapes, adapt to complex environments, and enhance safety in human-robot interactions [7][8][9][10][11]. Soft robots are widely utilized in the medical sector, especially for rehabilitation exercises, assisted grasping, and artificial organs [12]. To replicate and augment human organ functions, various bionic limbs and organs have been created, including soft robotic sleeve [13], soft stomach [14][15], and soft esophagi [16][17]. However, most robots designed to simulate the esophagus and stomach in the digestive system tend to overlook the simulation of the rectum, which is the terminal section of the digestive system, due to the associated shame and stigma (Fig. 1**A**).

Bowel control, particularly regarding fecal incontinence, can impact individuals across all ages and genders to varying extents [18][19][20][21]. Addressing this issue without shame is crucial, as effective treatments are available. People often take bowel control for granted, not fully appreciating the complex physiological processes involved. Bowel control depends on several factors, many of which can be modified. Improving bowel control can be approached through three primary aspects: enhancing the strength of the anal sphincter muscles, optimizing stool consistency, and managing anxiety, which can exacerbate symptoms. The anal sphincters, comprising two concentric muscle rings encircling the anus, are crucial for maintaining continence (Fig. 1**B**). The internal anal sphincter, forming the inner ring, remains contracted involuntarily except during defecation. The external anal sphincter, which constitutes the outer ring, is under voluntary control, allowing individuals to contract it for improved closure. These muscles work in unison to maintain continence. When stool enters the rectum, the internal sphincter relaxes, permitting the passage of stool into the upper anal canal. As depicted in Fig. 1**C** peristaltic movements (muscular contractions) in the rectum play a critical role in expelling stool from the body.

In recent decades, researchers have examined the rectum and anal canal from a biological perspective. F.S.P. Regadas et al. studied the anatomy of the anal canal using anorectal ultrasonography [22]. H. Kenngott et al. developed an open-source surgical evaluation and training tool called openHELP (Heidelberg laparoscopy phantom) [23].

M.B. Christensen measured the tensile properties of human and porcine rectal and sigmoid colon tissue [24]. N. Horvat et al. explored MRI for rectal cancer, focusing on tumor staging, imaging techniques, and management [25]. Additionally, other researchers have developed devices to replicate the functions of the rectum and anal canal. For instance, A. Doll et al. created a high-performance bidirectional micropump for a novel artificial sphincter system [26] . W.E. Stokes et al. proposed a biomechanical model of the human defecatory system to study continence mechanisms [27]. Y. Luo et al. designed SMA artificial anal sphincters [28]. K. Tokoro et al. developed a robotic defecation simulator [29]. E. Fattorini et al. discussed innovations and prospects for fecal incontinence treatment using artificial muscle devices [30]. L. Maréchal et al. utilized pneumatic and cable-driven mechanisms to model anal sphincter tone [31]. However, most of these devices are rigid, making it difficult to simulate the flexible nature of the rectum. Their complex drive structures and lack of physiological functions limit their widespread application.

To address this issue, we proposed the development of soft CMAs and soft rectum models to simulate the defecation process. Although various types of CMAs have been developed [32][33][34][35], they are challenging to attach to rectum models due to their complex shapes. In this study, we used the silicon rubber to manufacture both the rectum and the actuators, enabling perfect integration and attachment. Additionally, we synthesized a simulated stool using sodium alginate and calcium chloride. By emulating the occlusion-open mechanism of the anal sphincters and the peristaltic movement of the rectum using CMAs, we successfully replicated the defecation process. This system is expected to explore the complex physiological processes involved in bowel control and serve as a training model for inexperienced nurses, providing a hands-on approach to mastering the manual extraction of stool. Because it simulates the process of anal closure and rectal peristalsis, it can better assist beginners in understanding the defecation system from an educational perspective.

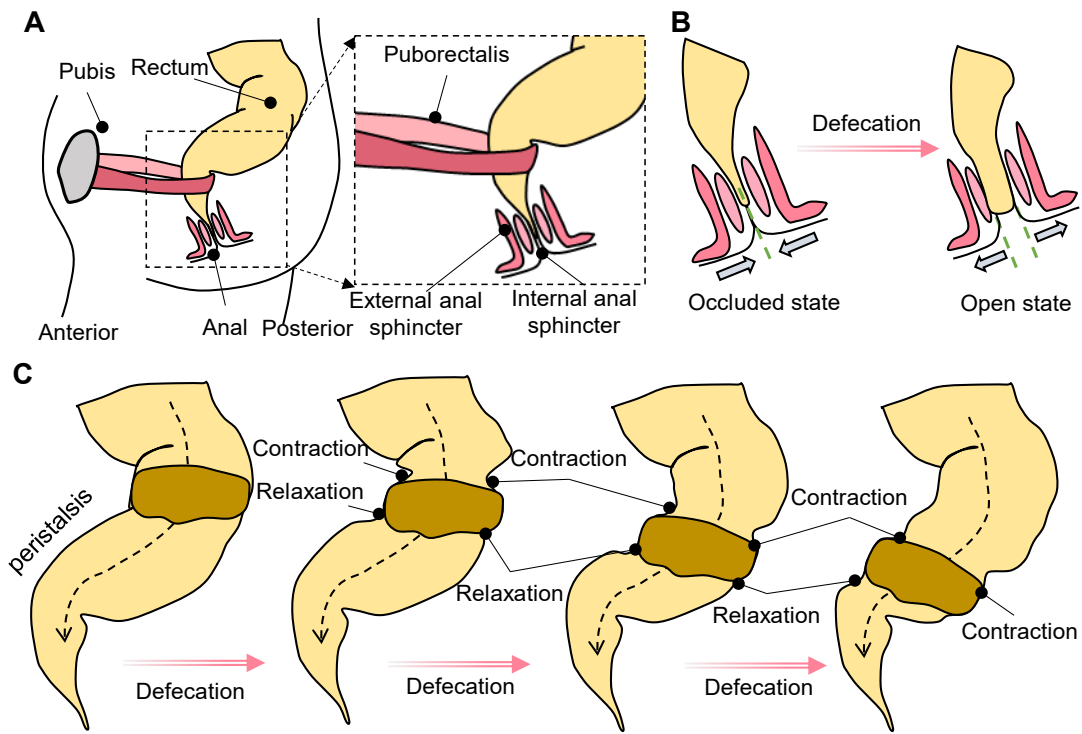

Fig. 1 **A:** An overview of the rectal system, encompassing the rectum, pubis, and anal sphincters. **B:** Mechanism of anal sphincter actuation. **C:** The defecation process: a detailed sequence of stages, from the resting state to the final expulsion of stool.

## 2. Materials and methods

### 2.1 Materials

The rectum model and CMAs are made of silicone elastomer (Ecoflex 00-30), with their molds made of PLA (polylactic acid). The stage in the system is also made of PLA. The CMAs are covered by PLA or PET (polyethylene terephthalate) sheets. The flush is made from Formlabs white resin cartridge v4. The simulated stools are synthesized using a sodium alginate solution and a calcium chloride solution. The pneumatic supply system consists of an air compressor (SR-L30MPT-01), a combined pressure controller and filter unit (FRF300-03-MD), a controller (Arduino Mega), and B005E1-PS on/off valves (Koganei Co., Ltd). Tubes (UB0425-20-BU) and adaptors (MPUC-4) are purchased from Pisco Company.

### 2.2 Mechanism and design

The normal human rectum possesses several functions and can perform basic defecation in response to various stool forms. In our study, we initially focused on modeling the function of anal closure for the expulsion of pellet-like stool (Fig. 2**A**), as this represents the fundamental function of the rectum and anus. For this purpose, we implemented three pneumatically driven actuators (CMAs) mounted on the rectal model. These actuators can transition from relaxation to fractionation due to an external gas actuation process. By controlling the movement of these actuators, we can compress the stool within the rectum until it is extruded. The area shrinkage of the actuators is crucial throughout this process, with a target value of nearly 1, and we will explore this aspect in subsequent sections. (Fig. 2**B**). In our design, we considered two different modes of operation. Pattern-1 allows actuator A1 to operate for an extended period, A2 for a shorter duration, and A3 for the final extrusion motion (Fig. 2**C**). In contrast, pattern-2 enables A1, A2, and A3 to operate for the same duration but in a sequential manner (Fig. 2**D**). The key distinction between Mode A and Mode B lies in the duration for which A1, A2, and A3 interact with the stool.

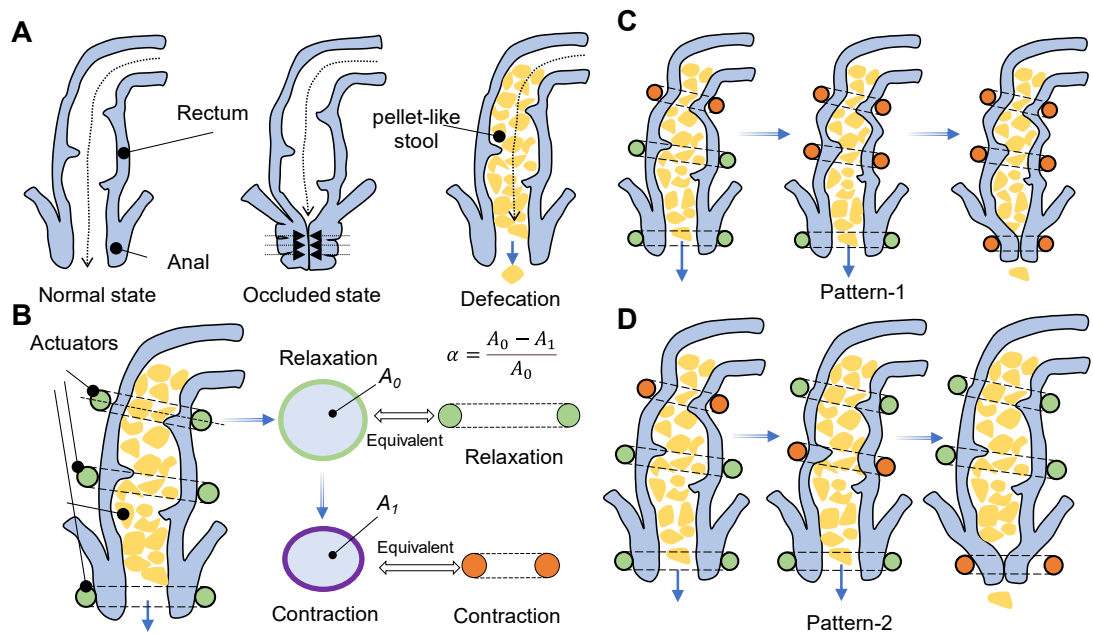

Fig. 2 **Principal Design and 3D Modeling: A:** The rectum in its normal state, the state with the anus closed, and during defecation. **B:** Three CMAs (Contractile Mechanism Actuators) are installed on the lateral side of the rectum, enabling contraction functions. **C:** Defecation in pattern-1 driven by CMAs **D:** Defecation in pattern-2 driven by CMAs: Sequential.

The rectum is the final section of the large intestine, located between the sigmoid colon and the anus. Its main physical movement is peristaltic contraction, which facilitates the storage and expulsion of feces. We created a 3D model of the rectum, with its critical dimensions illustrated in Fig. 3**A**. The curved surfaces of this model are intricate, particularly in the midsection of the rectum mold, which features multiple dimensions, complex torsion, and deformation structures. We highlight key dimensions such as radius, length, and diameter, with specific values provided (e.g., radius: 35 mm, length: 164 mm). Besides, In the rectum, longitudinal muscles promote the movement of feces toward the anus by muscle shortening, while circular muscles contract radially to compress and squeeze the feces. Smooth muscles are primarily responsible for the final expulsion of feces from the rectum. Through tonic contractions, the smooth muscles press the contents toward the anal canal, and the anal sphincters regulate the expulsion via peristaltic contractions of the circular muscles. Longitudinal muscles primarily control the relaxation and contraction of the rectum, coordinating with the smooth and circular muscles to complete the storage and expulsion of feces. The peristaltic contraction force and rate of the rectum are crucial for the effective expulsion of feces. To simulate this process, we designed three circular muscle actuators (CMAs), designated as A1, A2, and A3, which are sequentially connected to surround the rectal body and canal. These actuators mimic the natural contraction and peristalsis of the rectum. Fig. 3**B** illustrates the setup where air compressors connect to the rectum, employing a control strategy to simulate the contraction and peristalsis of the rectum. By adjusting the air pressure, we can precisely control the contraction patterns of the CMAs, thereby replicating the natural movements of the rectal muscles. This design aims to provide an accurate and functional model for studying the biomechanics of rectal movements and their role in fecal expulsion. Fig. 3**C** specifically illustrates the dimensions of A1, A2, and A3, including key parameters such as inner and outer diameters, and height. We can observe that the inner and outer diameters of A2 and A3 are 44 mm and 66 mm, respectively, while those of A1 are 27 mm and 53 mm. Fig. 3**D** showcases different aspects of the design, including various types of rings and components labeled as Type-I, Type-II, and Type-III, with different material layers like elastomer, PET paper, and PLA. We will evaluate their performance in the following sections.

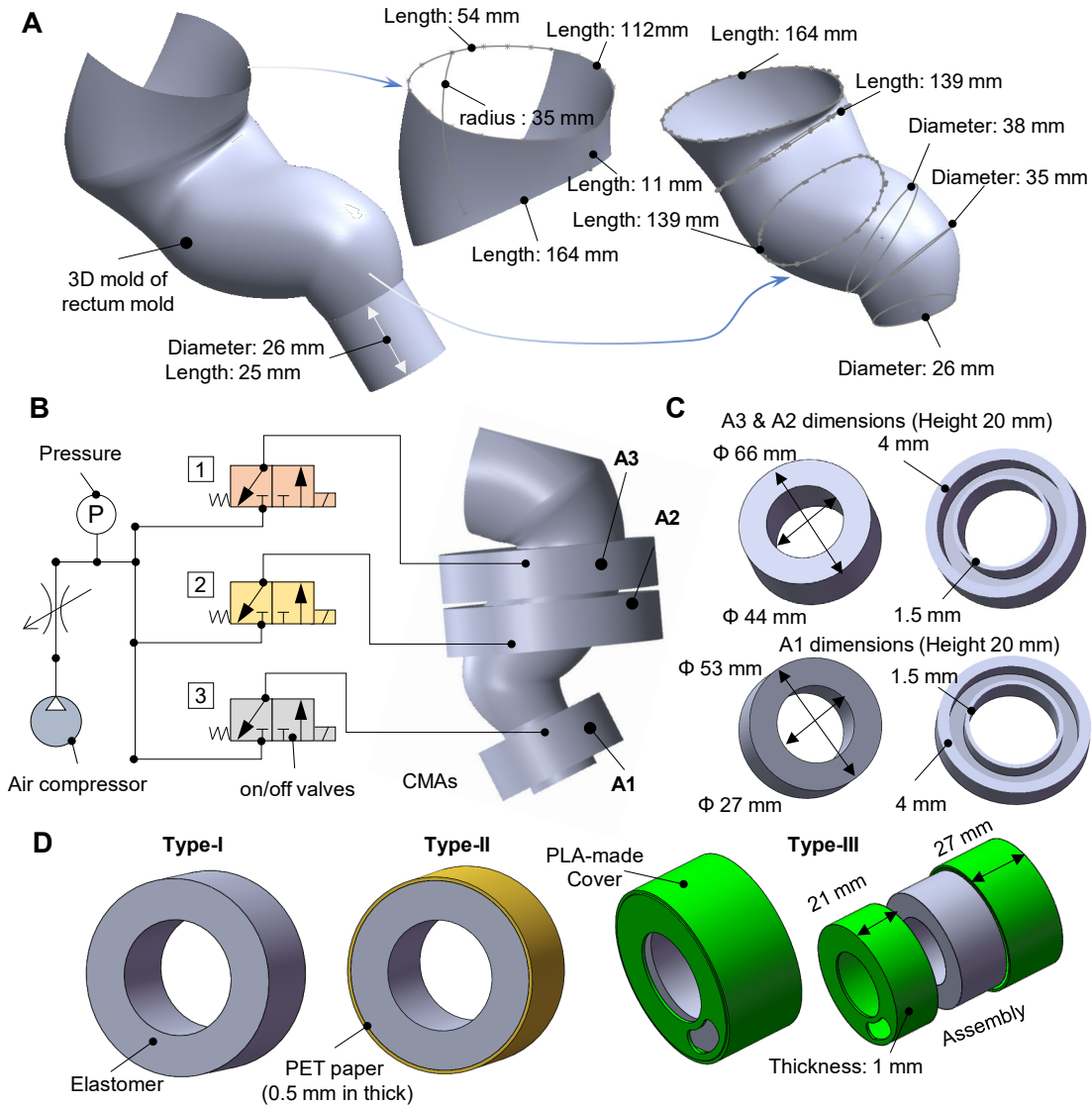

Fig. 3 Design of the Rectum Model and CMAs. **A:** 3D modeling of the rectum and its important dimensions. **B:** Arrangement diagram of CMAs in the model. **C:** Dimensions of CMAs. **D:** Design of three types of CMAs with different covers.

## 2.3 Fabrication of CMAs and rectum model

To fabricate these CMAs, we utilized 3D printing technology (RAISE3D Pro2) to create their molds, and then processed the CMAs using a replication molding technique. The ratio of Ecoflex part A to part B is 1:1. The mixture is stirred at a speed of 3000 rpm and subjected to a degassing process to obtain a homogeneous liquid mixture. The mixture is then poured into the prepared molds and cured in an oven at 60°C for 1 hour to produce the semi-finished products (Fig. 4**A**(i)). To facilitate demolding, Primer No. 4 (Shin-Etsu Chemical Co., Ltd.) was applied to the molds prior to sealing. Additionally, we used a punch to drill holes to ensure proper external connections. We designed and manufactured another cover mold, then poured Ecoflex 00-30 into it. After waiting for 5 minutes at room temperature, the punched semi-finished product was partially embedded in the cover, which was then placed in a constant temperature chamber for crosslinking (Fig. 4**A**(ii)). The photos of molds and fabricated CMAs without covers are shown in Fig. 4**A**(iii). We prepared several stacked actuators covered by PET papers to test the performance of CMAs (Fig. 4**B**).

Considering the fabrication of rectum model, the processing method is relatively simple, involving 3D printing to create a model, followed by injecting liquid Ecoflex 00-30. Firstly, we printed the complex model using a 3D printer, which includes the inner mold, bottom mold, and upper mold (Fig. 4**C(i)**). The bottom and upper molds were then assembled to form the mold assembly (Fig. 4**C(ii)**). After assembling, we inserted the inner mold into the prepared assembly (Fig. 4**C(iii)**). To prevent leakage, we reinforced the entire setup with clamps and then poured the prepared liquid Ecoflex 00-30 into the mold (Fig. 4**C(iv)**). The assembly was then placed in an oven at a constant temperature of approximately 60°C to cure. After curing, we removed the mold from the oven. Once the Ecoflex had solidified, we separated the molds to reveal the formed rectum model. In the final step, after demolding, we obtained the complete rectum model (Fig. 4**v**). This method ensures a precise and consistent fabrication process, resulting in a highly detailed and accurate rectum model suitable for further experimental testing. The use of Ecoflex 00-30 provides the model with the necessary flexibility and durability, closely mimicking the properties of human tissue.

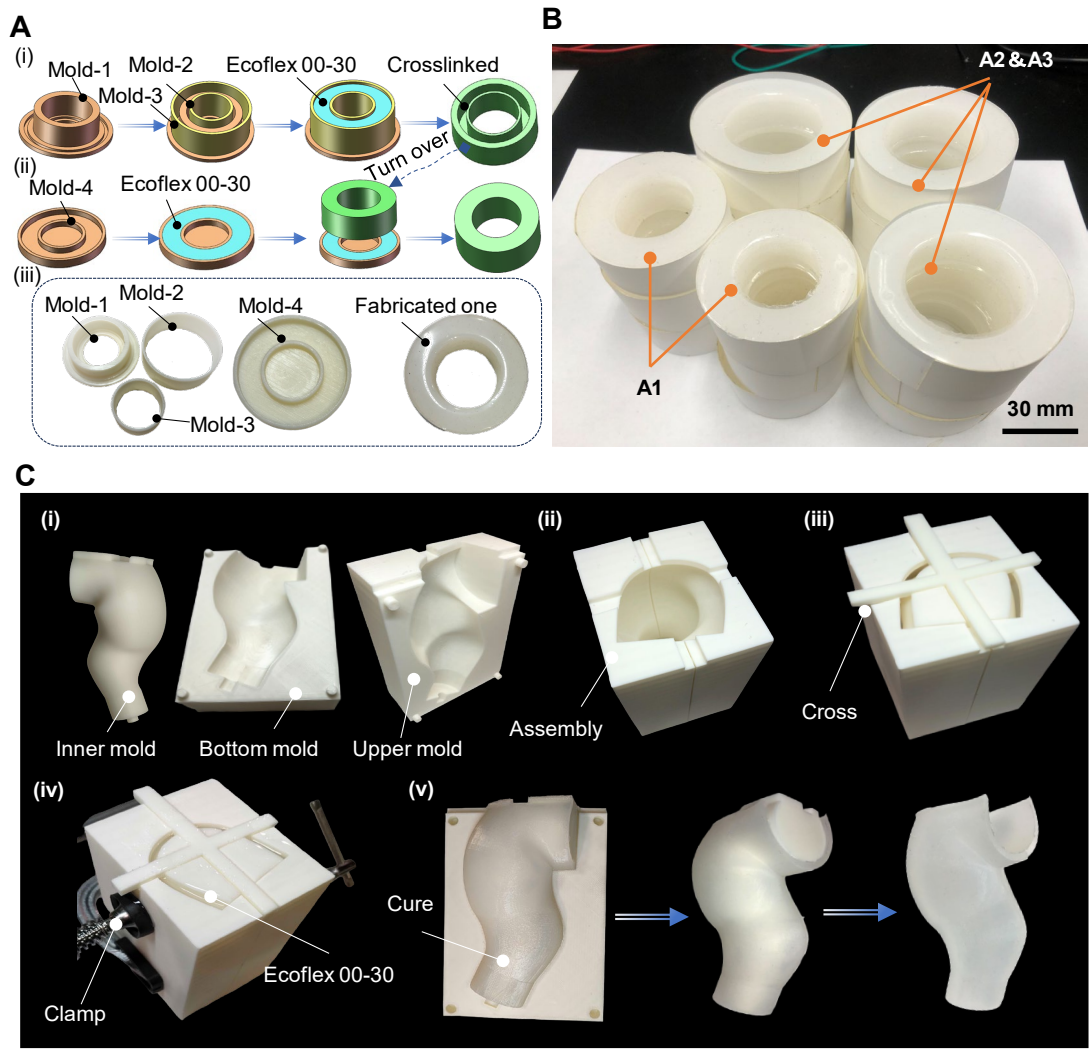

Fig. 4 Fabrication Process of CMAs and rectum model. **A:** (i) Fabrication of crosslinked base. (ii) Seal the base using the cover. (iii) Optical images of the molds used. **B:** Photographs of the CMAs. **C:** (i) Components of molds: inner mold, bottom mold, and upper mold. (ii) Assembly of the three molds. (iii) Fixation of the inner mold using the cross. (iv) Clamping of the molds and pouring of Ecoflex 00-30. (v) Removal of the upper mold and extraction of the rectum model.

## 3. Results

### 3.1 Performance of three types of actuators

Three different types of ring-shaped actuators were prepared to investigate their performance characteristics. The experimental setup for area contraction ratios and generated pressure are shown in Fig.S1. The first type involves soft actuators without any external cover, referred to as Type-I (Fig. 5). To mitigate the expansion effects on the outer and lateral sides of the soft actuators, we proposed two additional types: Type

II, which uses PET paper for wrapping, and Type III, which employs a 3D-printed rigid frame (Fig. 5**A**). At an applied pressure of 20 kPa, the inner membranes of Type II and Type III actuators exhibit noticeable expansion (Fig. 5**A**). For Type-I actuators, the deformation of the inner membrane as the pressure varies can be seen in Fig. 5**B**. It can be observed that the inner expansion of this soft actuator is not very uniform due to the difficulty in achieving a completely consistent wall thickness of the inner ring during the manufacturing process. Even with a uniform wall thickness, a bulge like that shown in the figure can form under pressure. It is evident that at an output pressure of 20 kPa, this bulge can effectively seal the inner ring hole. We analyzed and tested the area contraction of these actuators through simulations and experiments (Fig. 5**C**). To calculate the area contraction ratios, we used free software (ImageJ) to analyze the area changes based on the top-view images when the actuators are inflated or deflated. We evaluated the expansion of these ring-shaped actuators under different pressures. The area contraction ratio of the actuators increases with the increase in pressure (Fig. 5**D & E**). It is observed that Type I, Type II, and Type III actuators all display a contraction ratio of nearly 1 at 20 kPa.

Additionally, we tested the pressure generated on the inner side of the soft actuators of three different types and two different diameters. Regardless of the diameter sizes, Type III consistently produces the highest pressure, followed by Type II and Type I. From the simulation perspective, our results align well with the experimental data. Our simulation model is based on the dimensions after processing. This is because the precision of our mold is not very high, resulting in some parts of the final product being thicker while others are thinner. Therefore, we measured their approximate range and designed the same structure during the modeling process. Our simulation process is static, where we apply a pressure of 10 kPa to the inner cavity and observe its deformation. We then calculate the compressed area (top view) based on the deformed image, which allows us to obtain the area contraction ratios. As for the material settings, we set the soft material properties, such as density and Young's modulus, to be consistent with silicone (Ecoflex 00-30)

Moreover, we measured the internal pressure generated by the actuators when the input pressure was set to 10 kPa. The results indicate that the values of generated pressure follow the order: Type III, Type II, and Type I. It is also evident that actuators of Type A1 can generate higher forces compared to Type A2 and Type A3 actuators. Fig. 5**G** presents the deformation of Type-I actuators under different input pressures

($P_{in}$ = 0 kPa, 10 kPa, 12 kPa). This demonstrates how the actuator's deformation varies with increasing input pressure, highlighting the performance and mechanical behavior of the different types of actuators. These findings underscore the effectiveness of Type III actuators in achieving superior area contraction and pressure generation, making them a promising candidate for applications requiring precise control and high performance in soft robotics and related fields.

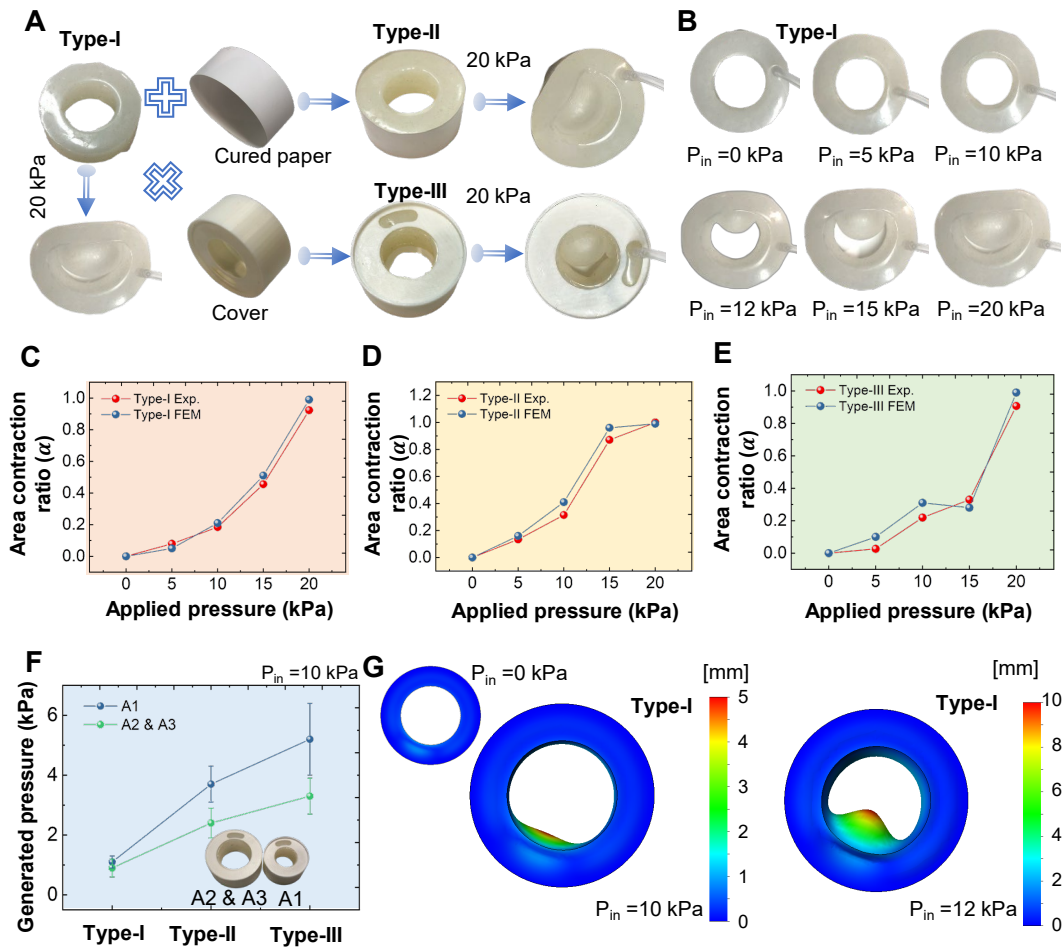

Fig. 5 Design and Comparison of Different Types of CMAs. **A:** Preparation and configuration of Type-I, Type-II, and Type-III structures. **B:** Optical images of Type-I actuators under varying internal pressures from 0 kPa to 20 kPa. **C:** Area contraction ratio for Type-I structure under different applied pressures. **D:** Area contraction ratio for Type-II structure under different applied pressures. **E:** Area contraction ratio for Type-III structure under different applied pressures. **F:** Comparison of generated pressure for different types of actuators with two designed patterns at 10 kPa internal pressure. **G:** FEM simulation results showing deformation in the Type-I structure at 0 kPa, 10 kPa, and 12 kPa internal pressures.

### 3.2 Preparation of simulated stool

Based on the Bristol Stool Chart, it categorizes stool into seven types, ranging from constipation to diarrhea. In this study, we primarily tested the defecating process using the prepared simulated feces, specifically small and round feces (Fig. 6**B**). To synthesize the simulated feces, we firstly dissolve 2 grams of sodium alginate powder in 98 grams of water to prepare a 2% (w/w) sodium alginate solution. Then, we stir the mixture continuously for one hour using an overhead stirrer at a speed of 1000 revolutions per minute to ensure complete dissolution. Before the experiment, we allow the sodium alginate solution to stand at room temperature (24.9 ± 0.5 °C) for 24 hours to eliminate any bubbles generated during stirring. As for the calcium chloride solution, we add the calcium chloride ($CaCl_2$) powder to a separate beaker of distilled water and then stir the solution until the calcium chloride is fully dissolved to achieve target concentrations of 0.15 M. Under this condition, we found that ratio of sodium alginate solution and calcium chloride solution is not prone to dry out. Next, we placed the syringe tip 7-8 cm above the gelling solution, carefully add the sodium alginate solution drop by drop into the calcium chloride solution using the syringe. The beads are manufactured and processed to fit within a length range of 15-20 mm and a width range of 5-12 mm (Fig. 6**A**). After the desired amount of gel beads is formed, we allow the gel beads to sit in the calcium chloride solution for a few minutes to ensure complete gelation. Finally, we rinse the gel beads with distilled water to remove any excess calcium chloride. To prevent the alginate gel from drying out, we store the gel beads in a sealed container, like in a refrigerator, to minimize exposure to air. This would help maintain the gel's moisture content and ensure consistent flow properties.

### 3.3 Controlling patterns

In this study, we employed a pneumatic supply system as the input method for the CMAs (Fig. 6**B**). The air compressor serves as the primary source of compressed air, which is regulated by the pressure controller and filter unit to ensure a consistent and clean air supply. The Arduino Mega controller is used to manage the operation of the pneumatic system, providing precise control over the actuation sequences. In the source code, we set the ON state for each of A3, A2, and A1 to 3 seconds and the OFF state to 1 second respectively (Pattern-1). For Pattern 2, the corresponding ON and OFF times are both 1 second. To achieve sequential operation, we adjusted the sequence through a for loop.

The B005E1-PS on/off valves are utilized to control the flow of air to the actuators, enabling the desired on/off control patterns. For the pneumatic connections, we used tubing with an inner diameter of 2 mm and an outer diameter of 4 mm. This choice of tubing ensures that the air flow is efficiently directed to the actuators, minimizing losses, and maintaining the desired pressure levels. By integrating these components, we achieved a reliable and effective pneumatic supply system capable of delivering precise control inputs to the CMAs.

We employed different control methods to simulate the defecation process, (Fig. 6**C & D**). These figures depict the varying control strategies applied to the actuators A3, A2, and A1. In Fig. 6**C**, the control times for A3, A2, and A1 within the cycle are set to 3 seconds. This implies that during each cycle, A3, A2, and A1 can influence the simulation separately and continuously, providing a sustained and coordinated control over the simulated fecal matter. In contrast, Fig. 6**D** shows a different approach where A3, A2, and A1 are subjected to a control pattern involving a brief one-second on/off control over the rectum model. This results in shorter compression durations for each actuator, leading to a more intermittent and less continuous control of the simulation. By comparing the results of these experimental setups, it becomes evident that the first method, with longer and continuous control times, allows for a more rapid expulsion of the simulated feces compared to the second method, which utilizes shorter and intermittent control bursts.

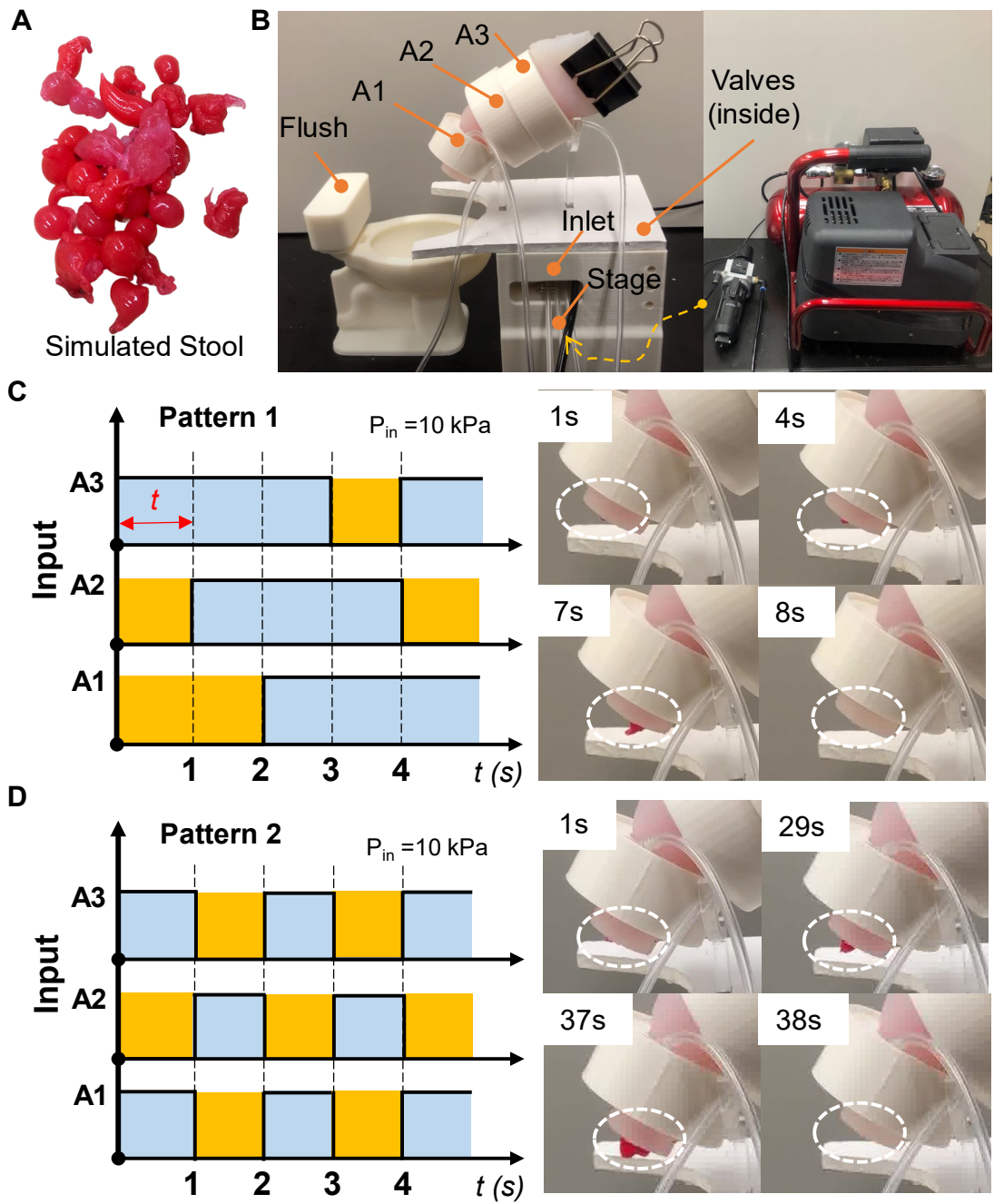

Fig. 6 Measuring systems, stimulated stool, and different control methods. **A:** Experimental setup with labeled components. **B:** Simulated stool used in the experiment. **C:** Input pattern and corresponding images showing the movement of the simulated stool over time (10 kPa). **D:** Alternative input pattern and corresponding images showing the movement of the simulated stool over time (10 kPa).

Additionally, we quantified the defecation velocity in relation to variations in working patterns (Fig. 7). The vertical axis represents the defecation velocity, ranging from 0.0 to 0.6 g/s, while the horizontal axis delineates the working patterns in a sequential manner. Four distinct working patterns were investigated, each defined as follows: (i) pattern-1 at t = 1s, (ii) pattern-2, (iii) pattern-1 at t = 2s, and (iv) pattern-1 at t = 3s. The parameter t is further elucidated in Fig. 6. The data reveal that the system subjected to pattern-1 exhibits the highest defecation velocity, averaging 0.42 g/s with a standard deviation of ±0.13 g/s. In contrast, pattern-2 results in a reduced velocity of 0.17 g/s, accompanied by a noticeable margin of error. These results substantiate the hypothesis that pattern-1 facilitates a more rapid defecation process. Furthermore, we analyzed the influence of the variable t on the defecation velocity within our system. It was observed that an increase in the value of t corresponds to a declining trend in defecation velocity, with measured values of 0.22 g/s and 0.11 g/s, respectively. All experiments were conducted under a constant control pressure of 10 kPa. These findings underscore the capability of the designed and developed CMAs to precisely replicate the defecation process. The results suggest that the sustained and coordinated control implemented in pattern-1 is more effective in simulating the natural defecation process, highlighting the potential application of these CMAs in biomedical research, particularly in studies aimed at simulating rectal function.

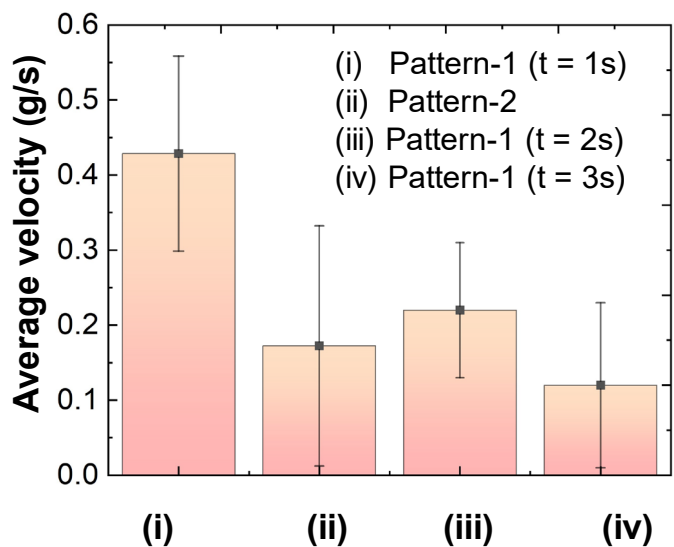

Fig. 7 Relationship between the defecation velocity and several working patterns.

## Conclusion

This study highlights the successful development and deployment of soft CMAs and rectum models that effectively simulate the defecation process. By incorporating Ecoflex 00-30, a highly flexible material, the research facilitated seamless integration of the actuators with the rectum models, closely emulating human tissue characteristics. The design and fabrication processes employed advanced techniques such as 3D printing and replication molding to ensure consistency. Three distinct types of CMAs were developed and evaluated, with the Type III actuators demonstrating superior performance in terms of area contraction and pressure generation. The integration of a pneumatic control system provided manipulation of the actuation sequences, effectively replicating the peristaltic movements in the rectum. Furthermore, the study explored two control patterns, finding that the second pattern, which utilizes all actuators to maintain a prolonged state, more effectively mimics the natural defecation process compared to a strategy involving short compression times. In the future, we will employ more precise control methods to simulate the defecation process.